\documentclass[lettersize,journal]{IEEEtran}
\ifCLASSINFOpdf
\else
\fi
\usepackage{algorithm}
\usepackage{algorithmic}
\usepackage{color}
\usepackage{graphicx}
\usepackage{amsmath}
\usepackage{comment}
\usepackage{diagbox}
\usepackage{xcolor}
\usepackage[english]{babel}
\usepackage[autostyle]{csquotes}
\usepackage{caption}
\begin{document}

\title{Filtering-out poor-quality images for data preparation}

\author{Roopdeep ~Kaur,
        Gour~ Karmakar,
        and~Muhammad ~Imran% <-this % stops a space
\thanks{R. Kaur, G. Karmakar and M. Imran are with the Institute of Innovation, Science and Sustainability, Federation University Australia
 e-mail: (roopdeepkaur@students.federation.edu.au, gour.karmakar@federation.edu.au, m.imran@federation.edu.au).}}

\maketitle

\begin{abstract}
    Filtering noise is a fundamental part of data preparation that enhances image quality for applications such as object segmentation, detection, and recognition. Various noise reduction techniques are proposed in the literature, including the use of median, Gaussian, and bilateral filters. Convolutional neural networks (CNNs) have gained popularity in image denoising owing to their ability to extract complex patterns and features from data. CNNs are highly adaptable, making them effective tools for various image-denoising tasks. One drawback of CNN-based techniques is that they require an appropriate training dataset and all images to be resized. Another notable drawback of all these filtering techniques is that they work  for certain types of environmental and camera noises. To bridge this research gap, in this paper, for the first time, instead of denoising, we propose an approach that filters out poor-quality images for various environmental and camera impacts. In our approach, quality is assessed using an image quality assessment metric and an optimum threshold is used to filter out poor-quality images. We also ensure that a sufficient number of images remain to develop the deep learning (DL) model. The results produced using real and simulated traffic and object recognition data demonstrate the performance supremacy of the proposed approach compared with the state-of-the-art approaches. The average recognition accuracy for our proposed approach is 93.8\% for the traffic sign recognition dataset and 84.9\% for the object recognition dataset. This indicates our model’s potential for real-life applications such as autonomous vehicles.
\end{abstract}

\begin{IEEEkeywords}
SSIM, image filtering, environmental impacts, camera impacts, convolutional neural networks.
\end{IEEEkeywords}

\IEEEpeerreviewmaketitle

\section{Introduction}
A digital image has significance in many fields such as smart health, smart agriculture, smart cities, wildlife conservation, surveillance and security, and industry \cite{nassereddine2024applications}. 
For example, smart agriculture applications may employ an Internet of Things (IoT) camera for capturing images to monitor crop health, detect pests, assess soil moisture, and manage irrigation systems. Farmers can use these images to make informed decisions about crop management, resulting in higher productivity and a reduction in resource consumption  \cite{kasera2024comprehensive}. 

However, the literature demonstrates that several environmental and camera distortion impacts lead to poor-quality image acquisition, which negatively affects the application performance because image-processing applications struggle to recognize or detect objects (e.g., traffic signs, pedestrians, vehicles) in those poor-quality images. Environmental impacts include snow, shadow (SH), rain, and darkness (DK), while lens blur and lens dirtiness are examples of camera impacts \cite{mittal2022survey}. For this, many image-denoising approaches \cite{jebur2023comprehensive, izadi2023image, kaur2023deep} have been proposed to remove noise while maintaining the quality of the image details, such as edges and sharpness. Image denoising is accomplished using a variety of standard filters such as Gaussian, median, Wiener filters \cite{monajati2019modified}. In real-world noisy images, however, these methods have some limitations and are not effective for other regular noise types that have not been specifically targeted as well as irregular noises. Therefore, to address this problem, there are many deep learning (DL)-based approaches employed for image denoising. The most prominent approaches are ResNet \cite{tai2017memnet} U-Net \cite{he2016deep}, and DenseNet \cite{ronneberger2015u}. However, the main limitation of these approaches is that they are suitable for a particular noise and do not work well for all types of noises. In addition, all of these approaches require training sets and resized images, which results in important information being lost.

Hence, tackling denoising issues still poses a challenge for researchers \cite{wu2023recent}. \textcolor{black}{To reduce the impact of environmental and camera noises on images, existing approaches employ mainly filtering techniques. However, the literature has demonstrated that the filtering approach has limited efficacy. This creates a compelling case for introducing an approach that can directly filter out poor-quality images from training and test datasets. This paper introduces an approach to filtering out poor-quality images affected by different types of environmental and camera distortion impacts. }

Our major contributions are as follows: 
\begin{enumerate}
\item For the first time, we propose a generic conceptual approach based on convolutional neural network (CNN) to filter out poor-quality images for data preparation so that the selected image dataset can be utilized by various applications. We employ an assessment metric to calculate the quality of the images and then filter out those images in which the quality is less than the threshold. The optimum value of the threshold is selected by maximizing the recognition accuracy of the validation set and ensuring sufficient images remain for training.
\item  We compare our proposed approach with state-of-the-art approaches. We use two datasets, Challenging Unreal and Real Environments for Traffic Sign Recognition (CURE-TSR) \cite{temel2019challenging} and Challenging Unreal and Real Environments for Object Recognition (CURE-OR) \cite{temel2018cure}, to calculate the traffic and object recognition accuracy. The average accuracy of our proposed approach is 93.8\% for CURE-TSR and 84.9\% for CURE-OR datasets, respectively, which is considerably different from the best-performing existing state-of-the-art approaches (77.4\% for Embedded Gaussian for CURE-OR and 77.5\% for CURE-TSR). 
\end{enumerate}
The paper is organized as follows. Section \ref{literature} presents the existing literature on various types of denoising techniques. The proposed filtering-out poor-quality images approach is presented in Section \ref{method}, while Section \ref{results} shows its performance in terms of recognition accuracy and computational time. Finally, concluding remarks and future works are provided in Section \ref{conclusions}.

\section{Literature Review} \label{literature}

\textcolor{black}{Image data understanding and preparation are the two main phases of the most commonly used data mining process model, namely the Cross-Industry Standard Process for Data Mining (CRISP-DM) model \cite{dinh2020survey}. The cleaning/filtering of image data is crucial step in data preparation that improves image quality and enhances model performance.
In image processing, images are denoised as a part of the image preparation process before they are fed into the model. The model learns relevant patterns more quickly and performs better when it is fed clean data.} Image denoising has been extensively investigated in various contexts such as medical imaging, surveillance and security, remote sensing, and forensic analysis \cite{bharati2021comparative}. However, most of the existing image denoising techniques are based on conventional approaches, which use standard filters \cite{10509824, 10990127} to improve the quality of images. Image denoising techniques can be categorized into two broad groups: (1) standard filter-based techniques, and (2) DL-based techniques. DL-based techniques are further classified into two categories: customized DL for denoising and embedded denoising filters in DL.

\subsection{Standard filter-based image denoising}
There are two types of filters for image denoising: linear and non-linear, each with its advantages and disadvantages. Linear filters include Gaussian, mean, and Wiener filters. Examples of non-linear filters are median and bilateral filters. Linear filters are advantageous owing to their fast processing times but have the disadvantage of not preserving edges effectively. Non-linear filters have the advantage of securing edges and the disadvantage of requiring more time for processing \cite{bharati2021comparative}.
Among these, the median filter is particularly recognized for its capability to remove salt and pepper (S \& P) noise, and is effective in clearing low-density impulse noise, thus improving edge detection. It employs a kernel of a positive odd integer size  \cite{kumar2022cartoonify}.  
In addition, the Gaussian filter is another effective tool for image denoising. Unlike other filters, it provides superior image smoothing and is built on the principles of Gaussian distribution  \cite{buades2005review}. 
Equation (1) represents the probability density function of a Gaussian distribution.

\begin{equation} 
P(x)=\frac{1}{\sqrt{2\pi\sigma^{2}}}e^{{-(x-\mu)^{2}}/{(2\sigma^{2})} }
\end{equation}

In this equation, $x$ represent the intensity of a pixel at a specific grey level within an image window, $\mu$ denotes the average intensity of all pixels in the same window, and $\sigma$ signifies the standard deviation. 
%\cite{ahmad2019comparative}. 
The degree of smoothing in a Gaussian filter is influenced by the standard deviation $\sigma$. This filter computes the Gaussian weighted average based on the surrounding pixel neighborhood. It prioritizes spatial proximity in its calculations, ignoring pixel intensity similarities and whether a pixel is part of an edge, which may result in edge blurring. This can affect algorithms that rely on edge information, such as segmentation or object detection, by decreasing the sharpness and definition of object boundaries. A mean filter blurs an image by replacing each pixel with the average of its neighboring pixels. Despite its simplicity, it is effective in reducing noise. A Wiener filter is one of the most fundamental methodologies for reducing noise. Wiener filtering eliminates additive noise. The blurring of the image is also inverted  \cite{puri2017analysis}. However, this filter has limitations when applied to noisy images in the real world that have complex characteristics. Image denoising can also be accomplished using wavelets \cite{10409615, 10420512}.
However, it is effective only in dealing with additive white Gaussian noise (AWGN). Without proper adaptation, this denoising technique does not work well for non-stationary noise or impulse noise. To overcome the filter limitations mentioned above, there exist various DL-based denoising techniques that often outperform traditional filtering methods, as discussed below.

\subsection{DL-based techniques for image denoising}
Image denoising methods have recently developed rapidly based on DL \cite{han2018framing}, \textcolor{black}{\cite{10814968, 10480580, ZHANG2026104786}}. DL-based techniques are divided into two sections below.
\subsubsection{Customized DL techniques for image denoising }
 The CNN-based denoising techniques use a specific architecture comprising a large number of convolutional layers to effectively remove noise from images. Typical examples of this type of architecture include ResNet \cite{tai2017memnet}, U-Net \cite{he2016deep}, and DenseNet \cite{ronneberger2015u}. When CNNs are deep, a vanishing gradient problem may occur, but skipping connections between neighboring layers can alleviate this problem. In U-Net, feature mappings are concatenated from the first to last convolutional layer, and from the second to the second-last convolutional layer. Convolutional blocks (containing multiple convolutional layers) are connected using skip connections. DenseNet connects convolutional layers to one another, overcoming ResNet’s limitations of selectively discarding some layers of information. However, a limitation of DenseNet is its depth.

As a solution, \cite{LIU2022171} exploits wavelet decomposition independence to divide each convolutional layer into multiple subnetworks, thus overcoming this problem of deep networks. 
Using a CNN framework that reorganizes numerous convolutional layers, the authors provided a novel solution to the issue of vanishing gradients. A large denoising challenge was divided into several smaller, independent challenges using wavelet decomposition. For each issue, noise in the sub-band is eliminated in a particular manner, direction, and size. Using batch normalization and residual learning combined with a CNN, Zhang et al. \cite{zhang2017beyond} were able to filter out the noise. CNNs are trained using noisy image patches and noise mapping to achieve blind denoising \cite{zhang2018ffdnet}. To enhance the robustness of a denoiser, the authors used a dual network to extract complementary features from noisy images in \cite{tian2020image}. Denoising effects were enhanced by merging several channels, making it possible to extract prominent features \cite{peng2021proceedings}. As per \cite{wang2020proceedings}, blind denoising is a two-phase process. A subnetwork is initially used to estimate the noise. In terms of the second method, optimized methods embedded into a CNN are commonly used for image denoising. To achieve the choice  between denoising performance and efficiency, meta-optimizers and CNNs were combined \cite{alawode2021meta}. In addition, Bregman iteration algorithms are particularly effective in converting depth image inpainting into image denoising \cite{li2019learning}.\textcolor{black}{Ding et al. \cite{10086612} introduces an unsupervised unified image enhancement network for simultaneous dehazing and denoising. The framework integrates dehazing, denoising, and fusion modules, leveraging their correlation for improved performance. Experimental results show superior dehazing and noise removal, particularly on real-world hazy datasets.}
The methods based on CNNs that have been discussed above demonstrate how effective CNNs are at denoising images. However, denoising models based on CNNs may not generalize well to images with different types or levels of noise. To train CNNs, a large amount of labeled data is required. Although noise-free images are abundant, obtaining images with different types of regular and irregular noises for training can be challenging and expensive. Further, robust models must be trained using diverse datasets that capture different types and levels of noise.

\subsubsection{Embedded denoising filters in DL}
\textcolor{black}{The term ``embedded denoising filter” refers to the integration of denoising filters directly into neural network architectures as a separate layer of the network. A key goal of this approach is to enable models to learn denoising as an integral part of their feature extraction and processing process.}

The authors \cite{tian2020image} introduced a conceptual approach called batch-renormalization denoising network (BRDNet) for image denoising, which can extract a clean image directly from a noisy environment. Two networks are combined to increase the network’s width and thus obtain more features. As batch renormalization is incorporated into BRDNet, it can deal with internal covariate shifts and small minibatch problems. Dilated convolutions are exploited as a separate layer to extract more information for denoising tasks. Research findings have indicated that the BRDNet performs steadily across both artificially and naturally noisy images. Nonetheless, it is ineffective for more complex, real, noisy image denoising, such as low-light and blurred images. 
 Yang et al. \cite{9696315} introduced a new denoising technique to enhance the performance of intelligent robot welding. The technique uses an attention-dense convolutional block to extract and accumulate multi-scale feature maps. It incorporates a residual bidirectional Conv Long Short-Term Memory (ConvLSTM) block, designed to learn long-range spatial context from localized feature maps, which works as a denoising filter. However, similar to prior DL-based models, it necessitates image resizing, which could either strip away crucial data or introduce unnecessary information, thereby potentially impairing tasks such as event detection and object identification.

Another study documented in \cite{9779501} merges the strengths of the CNN and transformer networks to refine real image denoising processes—crafting a hybrid model known as the Transformer Encoder and Convolutional Decoder Network (TECDNet). This design applies radial basis functions (RBFs) to improve the model’s representation while employing residual CNNs in place of more complex transformers to reduce computational demand. This results in the achievement of leading-edge denoising results on real images with reduced processing resource requirements. However, being efficient in consuming fewer resources, its denoising performance is comparable with other models but not superior.
Further, this paper \cite{kaur2023impact} aims to investigate how denoising affects CNN performance by first filtering noise from images using traditional denoising methods before they are fed into the CNN model. The authors then embed an image-denoising filter, namely Gaussian or median filters, as a layer into the CNN model. Denoising approaches are evaluated using both CNN accuracy and PSNR distribution. The comparative study provides insight into whether denoising will be used in various CNN-based image analyses, such as autonomous driving, animal detection, and facial recognition. The results of these traditional and embedded denoising methods require improvement in terms of recognition accuracy because only one filter is used as a denoising layer. Therefore, more filters can be added as a denoising layer.
All the above-mentioned techniques have the drawback of fixed size image usage, requiring all images to be resized, resulting in the loss of important information. In addition, the main limitation of the embedded techniques is the integration of standard filters.

In conclusion, denoising algorithms perform well on certain types of noise or image content, but they have difficulty with images that exhibit different characteristics or variations. It is possible that they are not capable of generalizing across diverse datasets. Most studied denoising algorithms train networks for different noise strengths and noise types based on the use of different parameter values. Noise in practice can be dynamically influenced by many environmental factors in the scene and/or intrinsic parameters from the camera. It is therefore essential to improve the robustness of the model against various noise characteristics. Poor-quality images consume additional resources, including the central processing unit (CPU) and storage, and these images confuse or deteriorate the applications’ decision-making processes. This indicates that the removal of these poor-quality images in both training and test sets will result in more accurate decisions. This is the main focus of research in this paper.

\begin{figure*}[hbt!]
\centering
\includegraphics[height = 2.5 in, width=6.5 in]{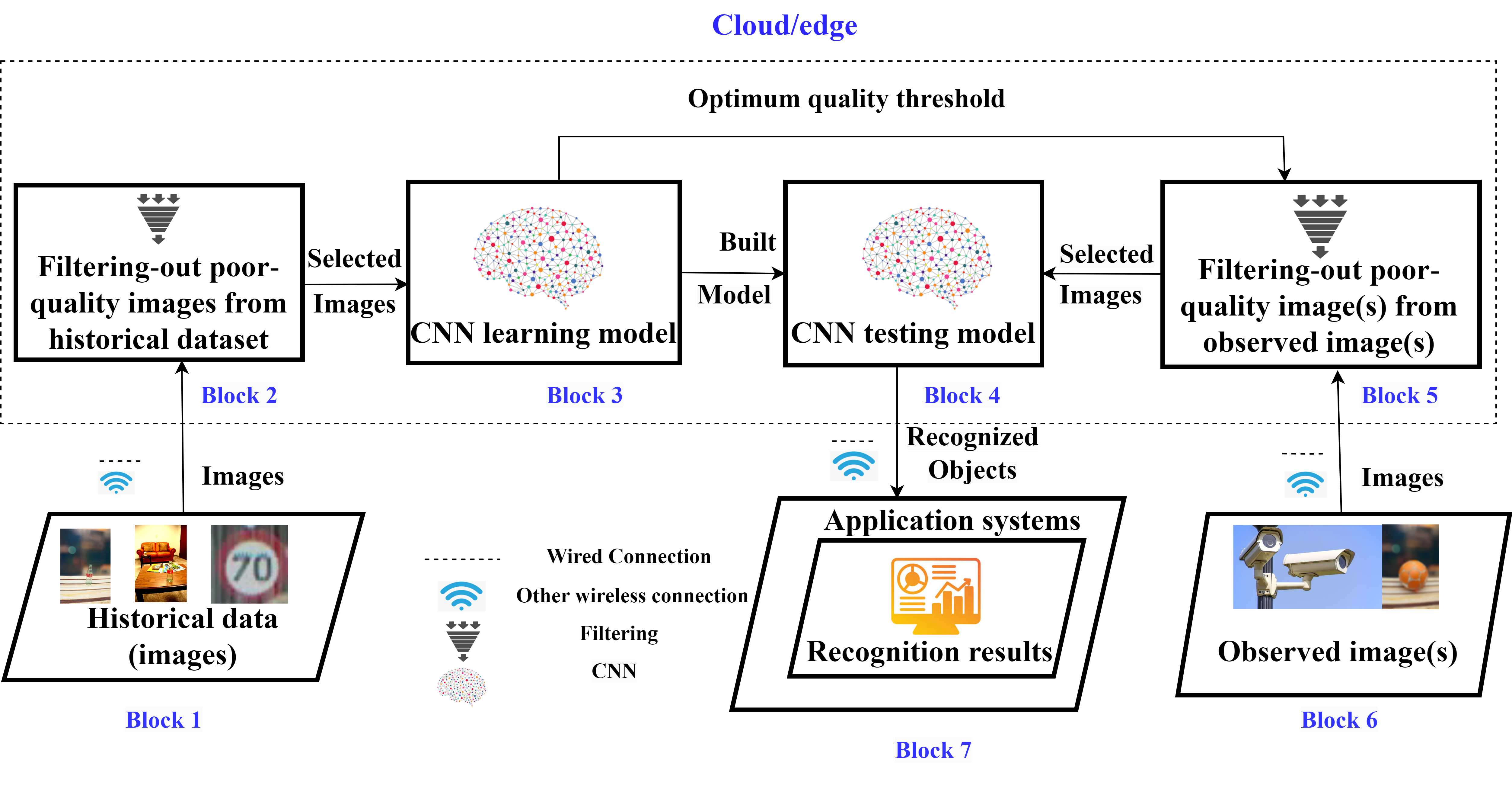}
\caption{Block diagram of our proposed poor-quality image filtering-out approach.}
\label{block1}
\end{figure*}  

\subsection{Research work motivation}
In practical applications, such as traffic sign detection systems in autonomous vehicles, numerous weather elements, such as rain, shadow, and snow, can degrade image quality by reducing contrast, introducing artefacts, and obscuring details. In addition, water droplets can cause blur and distortion on lenses. DL models incorporate standard filtering techniques for visual analysis. Since most denoising techniques are developed exploiting features of general noises, denoising techniques do not work well for all types of impacts. It is not possible to encode the characteristics of all types of effects using DL-based denoising techniques. To the best of our knowledge, there is no single method that can effectively detect and reduce various environmental and camera impacts, such as rain, SH, DK, and lens blur. The purpose of denoising algorithms is to remove noise from images, but they also inadvertently remove important information because of the artefacts inherently introduced by the filtering techniques. Therefore, to improve the recognition/detection accuracy, only images with sufficient information and clarity are required to be kept in the dataset. This challenging research gap can be addressed by filtering out the poor-quality images directly rather than using image denoising techniques for various environmental and camera impacts. 

\section{Methodology} \label{method}
The following sections in this paper present our proposed approach for filtering out poor-quality images. Figure \ref{block1} represents the schematic diagram of our proposed approach. In this block diagram, first, the historical data is used to filter out poor-quality images using quality metrics. Once the poor-quality images are filtered out using the threshold value of the quality metric, the CNN model is employed to calculate the recognition accuracy.

\begin{enumerate}
    \item 1)	Block 1 signifies the historical dataset. For a variety of tasks, including object recognition, image classification, and segmentation, historical data is used to develop the machine learning model. To train DL models, historical data is used as the foundation. There are two parts to the development process: (i) creating a training model, and (ii) selecting the appropriate machine learning model structure using the validation dataset. Therefore, historical data have two parts: (i) training, and (ii) a validation dataset to train machine learning models. During the training process, validation sets are used to evaluate a DL model’s performance. Training models on diverse historical datasets allows them to generalize better and perform better on different test data.
    \item  Block 2 filters out poor-quality images from the historical dataset and Block 3 is a CNN learning model. 
   As mentioned in the motivation section, to improve recognition/detection accuracy, we need to filter out poor-quality images that can confuse a model in both the training and testing phases.
    Therefore, in Block 2, we filter out poor-quality images from the historical dataset used in developing a DL model.
    Equation (2) represents our filtering criterion for filtering poor-quality images from the dataset:
    \begin{equation}
     [B_{m}, Q_{t}]=D_{m}(h,\epsilon,t_{p})\label{equation1}
    \end{equation}
where, $D_{m}(h,\epsilon,t_{p})$ denotes a function which develops the learning model after filtering-out poor-quality images. It returns the built model $B_{m}$ and optimum image quality threshold $Q_{t}$ i.e., the threshold for which the learning model ($B_{m}$) maximises the recognition accuracy.

$h$ represents the historical dataset and $\epsilon$ represents a threshold to make sure that sufficient images are remaining after filtering-out the poor-quality images for developing the CNN training model. $t_{p}$ is percentage of training set. 
The function of $D_{m}(h,\epsilon,t_{p})$ is detailed in Algorithm \ref{1}, which explains the developing CNN model. First, we calculate the quality scores for all images in historical dataset $h$ \textcolor{black}{using an image quality assessment metric} in Steps 3-5 of Algorithm 1. \textcolor{black}{Any image quality assessment metric %\cite{wang2004image, liu2019novel, rublee2011orb, vora2010analysis, aziz2015cycling} 
can be used to filter out poor-quality images.} 
We filter out poor-quality images that are below the particular value of quality scores in Steps 9-13. We then calculate the \% of filtered images, which should be greater than or equal to $\epsilon$ images (Step 14).
 For the given $t_{p}$, we divide the filtered images into training and validation sets (Step 15). Next, we develop the training model using the training set and calculate the accuracy of the trained model using the validation set (Step 17). Finally, we check that, if the accuracy is maximum and the \% of filtered images is greater than $\epsilon$, the quality scores are set to the optimum image quality threshold ($Q_{t}$). We store the trained model in the built model (Step 21) to use it for testing purposes. It is used to recognize/detect different objects in images. We select the CNN model because it is the best model for image analysis, including object recognition, compared with filters and transformers \cite{temel2017cure}. The CNN model has very little complexity since it consists of a few convolutional layers, pooling layers, fully connected layers, and a softmax layer for generating the final output. The hyperparameters for the CNN model can be selected after tuning the training model and can be varied based on the complexity of the datasets. The hyperparameters, including the number of epochs and layers, can be selected, which minimizes training and validation loss.

 \begin{algorithm}[H]
\caption{Algorithm for developing CNN learning model}
\begin{algorithmic}[1] \label{1}
\REQUIRE historical\_dataset ($h$), $\epsilon$ represents minimum \% of remaining filtered\_images; training\_percentage ($t_{p}$)
\ENSURE built\_model, optimal quality threshold $Q_t$
\STATE $\text{$A_i$}=0$ 
\textcolor{black}{\STATE $\text{$Q_t$}=0$  \\
// calculate the quality of all historical images
\FOR{$\text{image in historical\_dataset}$}
    \STATE $\text{quality\_score}(\text{image}) = \text{cal\_quality\_score}(\text{image})$
\ENDFOR\\}
%// do the following tasks by varying image quality threshold between [0.1, 1]. Filter out the poor-quality images from the historical dataset
\STATE\textcolor{black}{$\text{filtered\_images}=\text{historical\_dataset}$\\
\STATE $\text{temp\_historical\_dataset}=\text{historical\_dataset}$
}
\FOR{$i = 0.1$ to $1$}
     \FOR{$\text{image in temp\_historical\_dataset}$}
         \IF{$\text{quality\_score}(\text{image}) < i$}
             \STATE remove\_image\_from\_filtered\_images
         \ENDIF
    \ENDFOR\\
    // calculate the percentage of filtered images
    \STATE $\text{filtered\_images\%} = (\#\text{filtered\_images} / $ \\ \#\text{historical\_dataset}) 
    $\times 100$\\
    %// Divide filtered images into training and validation sets
    \STATE [training\_set, validation\_set]=divide (filtered\_images, training\_percentage)\\
     %// Develop a training model using the training set
    \STATE [trained\_model]=train(training\_set, model, criterion, optimizer, \#epoch) \\
     %// Get  the accuracy A of the trained model using the validation set
    \STATE [A]=evaluate (validation\_set, trained\_model, criterion)\\
    // find the optimal quality threshold and trained\_model 
    \IF{$A_i<A$ AND filtered\_images\% $\geq$ $\epsilon$}
        \STATE $A_i=A$ // keep the improved accuracy
        \STATE $Q_t=i$ // set optimum image quality threshold to i
        \STATE built\_model=trained\_model  
        %// store the trained\_model in built\_model for testing purposes
      %built\_model=trained\_model
    \ENDIF\\
     %// increase image quality threshold by 0.1
    \STATE $i=i+0.1$\\
    %\textcolor{black} {// $set temp\_historical\_dataset to filtered\_images$\\ 
$\text{temp\_historical\_dataset}=\text{filtered\_images}$
\ENDFOR
\end{algorithmic}
\end{algorithm}
\item Block 4 is the CNN testing model and Block 5 filters out poor-quality images from observed images. In Algorithm 2, first, we calculate the quality scores for all images in the test set (Steps 2-3). Second, we filter out poor-quality images from the test set using the optimum image quality threshold ($Q_{t}$) calculated in Algorithm \ref{1}. The filtering-out technique will decide whether filtering is required or not. \textcolor{black}{The trained model is employed to test filtered observed images in Block 4.}
\item Block 6 contains observed images. These images are captured from the camera affected by environmental and camera impacts and will be stored in a test set, which will be further utilized for filtering out poor-quality images.
\item Block 7 represents recognition results. We calculate the maximum recognition accuracy $A$ and computational time $C$ with the CNN testing model which is given in Block 4. The CNN testing model is a model that will be applied in image processing applications.

\end{enumerate}

\begin{algorithm}
\caption{Algorithm for testing CNN model} 
\begin{algorithmic}[1] \label{2}
\REQUIRE test\_set, built\_model $B_{m}$, optimal quality threshold $Q_t$ 
%$\epsilon$ represents minimum \% of remaining filtered\_images; training\_percentage
\ENSURE recognition accuracy A, computational time C 
    \STATE $\text{filtered\_test\_images}=\text{test\_set}$
    \textcolor{black}{\FOR{$\text{image in test\_set}$}
    \STATE $\text{quality\_score}(\text{image}) = \text{cal\_quality\_score}(\text{image})$
        \IF{$\text{quality\_score}(\text{image}) <$ $Q_t$}
        \STATE remove\_image\_from\_filtered\_test\_images
        \ENDIF
\ENDFOR \\
// get the test results using the filtered\_test\_images\\
\STATE [A, C] = evaluate (filtered\_test\_images, $B_{m}$, criterion) }   
\end{algorithmic}
\end{algorithm}

\section{Results and Discussions} \label{results}
This section covers experimental setup, datasets used in our experiments, CNN model structure, and finally, the results based on our experiments. In our experiments, we first determine the optimum quality threshold value based on a historical (validation) dataset. Next, we filter out poor-quality images from the dataset based on the optimum quality threshold value. Finally, we use the CNN model to calculate the traffic sign, object recognition accuracy, and computational time. To show the efficacy of our proposed filtering out approach, we compared our proposed approach with baseline approaches such as ``without denoising", ``embedded median", ``embedded Gaussian", ``traditional median", and ``traditional Gaussian" and demonstrated the efficacy of our results. %Baseline approaches" \cite{kaur2023impact}. 
%\textcolor{black}{preface please}

\subsection{Experimental setup}
Our experiments were conducted on an HP Zbook 15 G6 laptop with an Intel Core i7 vPro 9th Gen processor and an NVIDIA Quadro T1000 GPU. The HP Zbook has a physical memory of 32 GB. Using the PyTorch library, we implemented our model using Python in Visual Studio Code software. 
\subsection{Datasets}\label{dataset}
The CURE-TSR \cite{temel2019challenging} and CURE-OR \cite{temel2018cure} datasets were used for filtering. These datasets contain both real and synthetic images embedded with noise generated from the environment and the camera. There are five different levels of noise in both datasets: (i) extreme less, (ii) less, (iii) moderate, (iv) high, and (v) extreme high. However, the percentage of noise level is not provided in either dataset. Among the different traffic sign types in CURE-TSR are speed limit, goods vehicles, no entry, priority to, no left, no stopping, no overtaking, no parking, bicycle, stop, hump, no right, parking and yield. For various environmental impacts such as rain, SH, DK, and snow, we use all traffic sign images, including those shown in Block 1 of Figure \ref{block1}. Camera impacts include lens blur and lens dirtiness, which are two of the most prominent impacts.

\begin{figure}
\includegraphics[width=3.5 in]{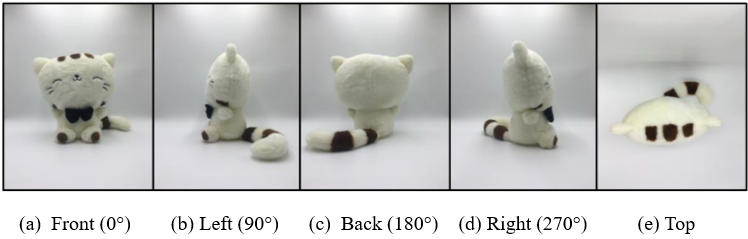}
\caption{Different orientations (stuffed animal) of the same object with the same background in CURE-OR dataset \cite{temel2018cure}.
}\label{orientation}
\end{figure}

\begin{figure}
\includegraphics[width=3.5 in]{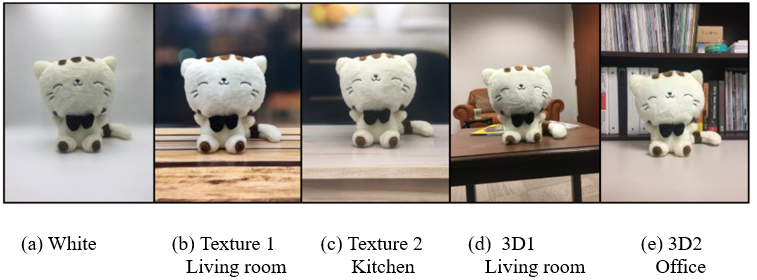}
\caption{Different backgrounds (stuffed animal) of the same object with the same orientation in CURE-OR dataset \cite{temel2018cure}.
}\label{background}
\end{figure}

CURE-OR comprises 23 toy categories, 10 personal item categories, 27 household item categories, 14 office item categories, 10 sports/entertainment categories, and 16 health categories. This dataset contains 100 classes of traffic signs, while CURE-TSR contains 14 classes. Five types of orientations of the same objects with the same background were used for the CURE-OR dataset, as shown in Figure 2. Images for lens dirtiness, lens blur, OE , S \& P noise, CT, and UE impacts were taken from the CURE-OR dataset, which includes all 100 object classes. As per the literature, complex/cluttered backgrounds make it difficult to recognize objects \cite{golcarenarenji2022machine}. To demonstrate the type of backgrounds used with the CURE-OR dataset images, Figure \ref{background} shows an example of a set of images with the various backgrounds. To test the accuracy of the traffic sign detection CNN model, we used 36458 training and 16670 test images. A total of 11220 training images and 3750 test images are used in the CURE-OR dataset. %\textcolor{black}{In Tables \ref{tab:freq1} and \ref{tab:freq3}, S \& P stands for S \& P, CT means CT, OE is overexposure and UE is underexposure.}. In Tables \ref{tab:freq mean} and \ref{tab:freq2}, SH means shadow and DK is DK.
We divided both datasets (CURE-TSR and CURE-OR) into two equal parts:\\ (i) Historical dataset: it is further divided into training (i.e., 80\%)  and validation (i.e., 20\%). The validation set was used to select the optimal quality threshold value and the epoch number.
(ii) Performance analysis dataset \textcolor{black}{(test dataset)}: The other half dataset is used for the performance analysis of recognition accuracy and \textcolor{black}{CPU} time. In the CURE-TSR dataset, the historical dataset contains the first 1-7, and the performance analysis dataset includes classes from 8-14 classes of traffic signs.
In the CURE-OR dataset, the first half dataset that is used as historical data contains all backgrounds of images of 100 types but includes only front (0$^{\circ}$), left (90$^{\circ}$), and half of the historical dataset of back (180$^{\circ}$) orientation as shown in Figure \ref{orientation}. The other half i.e., the performance analysis dataset includes half the dataset of back (180$^{\circ}$), and all right (270$^{\circ}$) and top oriented dataset.

\subsection{Description of the CNN model and its parameters}
Our CNN model consists of three convolutional layers, three fully connected layers and two max-pooling layers. We used a softmax classifier, which gives probabilities for each class label, as shown in Table \ref{tab:freq100}. 
 %Softmax classifiers were used in our CNN model to provide probabilities for each class label. 
 Based on \cite{temel2017cure}, we used 55 epochs for our CNN model. 
 A learning rate of 0.1 was used for our experiments. For our experiments, we used 256 batch sizes.

\begin{table}[hbt!]
\caption{Various parameters and their type/values for CNN.}
\label{tab:freq100}
\centering
\begin{tabular}{|c|c|c|c|c|c|}
\hline
{\textbf{Parameter}}  & {\textbf{Type/Value} }\\
\hline
\hline
Epochs & 55  \\
\hline
Learning rate & 0.1 \\
\hline
Batch size & 256 \\
\hline
 Classifer & Softmax\\
\hline
Activation function & ReLU\\
\hline
Max-pooling  & 2 \\
\hline
Convolutional layers  & 3\\
\hline
Fully connected layers & 3\\
\hline
\end{tabular}
\end{table}

 \subsection{Performance metrics}
 %Performance metrics are measurements used to evaluate the efficiency, effectiveness, and success of a system. 
 We \textcolor{black}{use} two metrics to measure the performance of the system which are given below.
 \paragraph{Recognition accuracy}
 The ability of a computer system to identify or classify objects, patterns, or features in an image or video stream is called ``recognition"; thus, it is assessed by recognition accuracy.
%Using the formula below, we calculate CNN's Top-1 recognition accuracy $\mathcal{X}$ for various traffic signs and objects.\\
%\begin{equation}
%\mathcal{X} = \mathcal{CP}/\mathcal{TP}
%\end{equation}\\ 
Recognition accuracy is the ratio of the number of correct predictions to the total number of predictions. %are represented by $\mathcal{CP}$ and $\mathcal{TP}$, respectively. 
%Tables \ref{tab:freq1} and \ref{tab:freq mean} present the Top-1 recognition accuracy for CURE-OR and CURE-TSR datasets.
\paragraph{Computational time}
Computational time analysis measures the time it takes for a computer system to perform a specific task, such as recognition or classification. As it affects overall system performance and efficiency, this is an important metric in computer vision and pattern recognition. Besides computational complexity analysis, computational time is also measured by CPU time. It is the testing time for recognizing the objects  and was measured in seconds for our experiments.
%and provided in Tables \ref{tab:freq3} and \ref{tab:freq2} for CURE-OR and CURE-TSR datasets. 
%Computational time is provided to perceive the computing resources required compared with the technical specifications of the machine used to implement the proposed approach suitable for a particular application. It is testing time for recognizing the objects.

\subsection{Experimental results for CURE-OR dataset}
As per the literature, the SSIM  is the most reliable image quality assessment metric used to assess image quality \cite{kaur2023evaluating}.
We utilized SSIM for filtering-out poor-quality images. Following are the experimental results for the CURE-OR dataset.

\begin{table}[hbt!]
\caption{Various impacts in CURE-OR and their \textcolor{black}{optimum image quality (SSIM)} threshold values. Note, OSTV represents optimum SSIM threshold value and RI is the remaining images. }
\label{tab:freq10}
\centering
\begin{tabular}{|c|c|c|c|c|c|}
\hline
{\textbf{Impact}}  & {\textbf{OSTV}} &{\textbf{Accuracy(\%)}} & {\textbf{RI(\%)}}\\
\hline
\hline
Lens blur & 0.9 & 88.4 & 20\\
\hline
Lens dirty & 0.1 & 93.6  & 100 \\
\hline
S \& P & 0.02 & 87.5 & 64 \\
\hline
CT & 0.1 & 97.5 &  100 \\
\hline
OE & 0.9 & 20.8  & 12 \\
\hline
UE  & 0.1 & 52.8 & 100 \\
\hline
\end{tabular}
\end{table}

\subsubsection{Estimating the threshold for filtering-out poor-quality images for CURE-OR dataset}
\begin{figure}
\includegraphics[width=3.5 in]{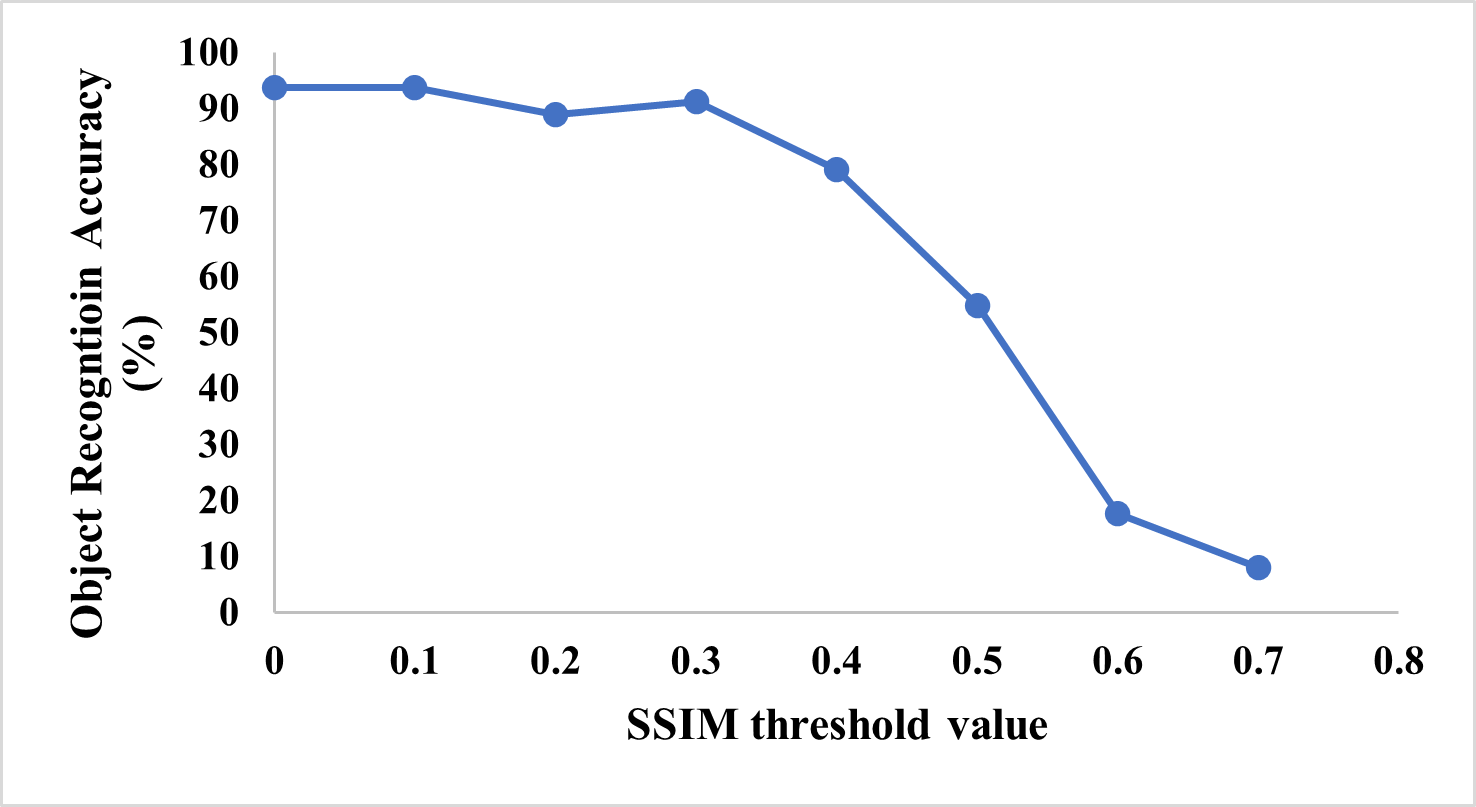}
\caption{Object recognition accuracy vs SSIM threshold value for lens dirty validation dataset of CURE-OR. Note, that SSIM threshold values were used for filtering-out poor-quality images from both training and validation datasets.}\label{lens dirty}
\end{figure}

Table \ref{tab:freq10} refers to different impacts in the CURE-OR dataset and the selected SSIM threshold values for filtering-out poor-quality images.\\ For example, how recognition accuracy varies with the SSIM threshold value is shown in Figure \ref{lens dirty}. For lens dirty, the recognition accuracy decreases with an increase in the SSIM value and it is maximum (93.6\%) at SSIM value = 0.1 \textcolor{black}{and the percentage of remaining images is 100}. Thus, SSIM value 0.1 is utilized as a filtering criterion for the performance analysis of recognition accuracy versus SSIM for lens dirty. Similarly, for CT and UE, the recognition accuracy is maximum at SSIM value = 0.1. For CT, the maximum recognition accuracy is 97.5\% and 52.8\% for UE. \textcolor{black}{In both impacts, the percentage of remaining images is 100.}
 In lens blur, the object recognition accuracy increases with the rise in SSIM value above 0.5. It reaches its maximum accuracy (88.4\%) at SSIM value = 0.9. \textcolor{black}{At this SSIM value, the percent of remaining images is 20 as shown in Table \ref{tab:freq10}.}
%\textcolor{black}{what about the percentage of images remaining??? write it for all optimal SSIM threshold values}. 
Therefore, we selected SSIM threshold value of 0.9 for filtering-out poor-quality images in lens blur from the historical dataset.

For S \& P, the recognition accuracy increases with an increase in SSIM until it reaches the value of 0.02 and then it gradually decreases. The maximum recognition accuracy is 87.5\% at SSIM value = 0.02, and the remaining images are 64\%. For OE, the recognition accuracy reaches its maximum value (20.8\%) at SSIM value = 0.9. Therefore, in all impacts covered in the CURE-OR dataset, the relevant SSIM values are selected based on their corresponding maximum recognition accuracies in the historical dataset.

\begin{table}[hbt!]
\caption{Object recognition \textcolor{black}{(OR)} accuracy (\%) of CNN in different environmental and camera impacts with various denoising approaches for CURE-OR.}
\label{tab:freq1}
\centering
\begin{tabular}{|p{1.4cm}|p{0.5cm}|p{0.5cm}|p{0.7cm}|p{0.5cm}|p{0.5cm}|p{0.5cm}|p{0.5cm}|}
%\begin{tabular}{|c|c|c|c|c|c|c|c|c|c|c|c|c|c|}
\hline
& \multicolumn{7}{|c|}{\textbf{OR accuracy for each impact (\%)}}   \\
 \hline
{\diagbox[width=\dimexpr \textwidth/13+2\tabcolsep\relax, height=1cm]{\textbf{Approach}}{\textbf{Impact}}} & {\textbf{Blur} }& {\textbf{Dirty}} & {\textbf{S \& P}} & {\textbf{CT}} & {\textbf{OE}} & {\textbf{UE}}  & {\textbf{Mean}}   \\
\hline
\hline
Without Denoising  & 89.3 &  \textbf{95.6} & 81.8 &  \textbf{96.6}  & 0.96 &  \textbf{98} & 77	\\
\hline
Embedded median & 85.12 & 	84.2 &	73.2	& 90.6	 & 0.96 & 	89.1 & 70.5
   \\
\hline
Embedded Gaussian & 91.5 &	94	& 85.3	& 95.3	& 0.96 &	97.7 & 77.4
  \\
\hline
Traditional median & 90.5 &	93.6	 & 84 &	95.6 &	0.96 &	97.4 & 77
  \\
\hline
Traditional Gaussian & 91.1	& 94.08 &	83.6 &	96 &	0.96 &	96 & 76.9
  \\
\hline
\textcolor{black}{Proposed} filtering-out & \textbf{95.2}	& \textbf{95.6}	& \textbf{87.6}	& \textbf{96.6}	& \textbf{36.6} &	\textbf{98} & \textbf{84.9} \\
\hline

\end{tabular}
\end{table}

\subsubsection{Performance analysis}
To analyse the performance of \textcolor{black}{the CNN model using our proposed filtering-out poor-quality images approach}, as alluded before, we chose recognition accuracy and computational time as the two metrics. Below are the results for recognition accuracy and computational time.
\paragraph{Recognition accuracy analysis}
We tested the traffic sign and object recognition accuracy of our filtering-out approach against approaches without denoising, traditional denoising and CNN-based embedded denoising.
The recognition accuracy of CURE-OR is mentioned in Table \ref{tab:freq1}. The table shows the comparison of object recognition accuracy with our proposed filtering-out approach and other denoising approaches (embedded and traditional) using median and Gaussian filters and no denoising.
To illustrate, in lens blur, \textcolor{black}{our proposed} filtering-out approach has maximum recognition accuracy i.e., 95.2\% in comparison with other denoising techniques. This is because the poor-quality images with an SSIM of less than $0.9$ are being filtered \textcolor{black}{out} in our proposed approach \textcolor{black}{(see Table \ref{tab:freq10})} rather than denoising images with median and Gaussian filter traditionally and in CNN-based embedded approach. The minimum recognition accuracy in lens blur is for the embedded median approach which is 85.12\%. \\
In lens dirty, filtering-out (95.6\%) and without denoising (95.6\%) approaches achieve a superior performance compared with other traditional and CNN-based embedded approaches. Various types of distortions and artefacts can be introduced into an image via dirty lenses, such as blurring, streaks, smudges, and lens flares. In contrast with typical noise or blur patterns that Gaussian and median filters are designed to handle, these artefacts are often more complex and irregular. Therefore, it is better to filter out poor-quality images with an SSIM of less than 0.1 rather than denoise images using median and Gaussian filters.\\
Similarly, for S \& P, the filtering-out approach has maximum recognition accuracy (i.e., 87.6\%) compared with other approaches.
In contrast, the filtering-out approach (96.6\%) and the approach without denoising work better compared with other approaches. In some cases, Gaussian and median filters can improve CT as a result of noise reduction or outlier removal, but they are not specialized tools used for this purpose. Therefore, \textcolor{black}{for CT, the approach filtering-out poor-quality images appears to be an effective vehicle.} Similar results can be seen in the UE as it has maximum recognition accuracy in filtering-out and without denoising approaches.
In OE, the performance of the filtering-out approach (36.6\%) is the best in contrast with other traditional and embedded approaches. OE typically results in loss of detail and clipping of pixel values in the highlights, which cannot be fully recovered using basic filtering techniques such as median and Gaussian filters. Since we have multiple exposures of the same scene, filtering-out overexposed images may still leave us with enough data from other exposures to generate a good result, as is reflected in Table \ref{tab:freq1}. \textcolor{black}{Overall, our proposed approach has maximum average accuracy for all impacts (i.e., 84.9\%) among the state-of-the-art approaches which demonstrates the effectiveness of our proposed approach.}

\begin{table}[hbt!]
\caption{Computational time (\textcolor{black}{CPU time in} seconds) per image for object recognition in different impacts with various denoising approaches for CURE-OR.}
\label{tab:freq3}
\centering
\begin{tabular}{|p{1.5cm}|p{0.5cm}|p{0.5cm}|p{0.7cm}|p{0.5cm}|p{0.5cm}|p{0.5cm}|p{0.5cm}|}
%\begin{tabular}{|c|c|c|c|c|c|c|c|c|c|c|c|c|c|}
\hline
& \multicolumn{6}{|c|}{\textbf{Computational time (seconds) for each impact}}\\
 \hline
\diagbox[width=\dimexpr \textwidth/12+2\tabcolsep\relax, height=0.8cm]{\textbf{Approach}}{\textbf{Impact}}  & {\textbf{Blur} }& {\textbf{Dirty}} & { \textbf{S \& P}} & {\textbf{CT}} & {\textbf{OE}} & {\textbf{UE}}  \\
\hline
\hline
Without denoising  & 2.98 & 3.27	 &	3.36 &	3.26 &	2.98 & \textbf{2.94} \\
\hline
Embedded median &3.12 & 3.41 & 3.47 & \textbf{2.96} & 3.04 & 3.10
\\
\hline
Embedded Gaussian & 3.21 & 3.37 & 3.88 & 3.42 & 3.22  & 3.36
\\
\hline
Traditional median & 3.18 & 3.27 & 3.43 & 3.20 & 3.19  & 3.17
\\
\hline
Traditional Gaussian & 3.17 & \textbf{3.24} & 3.31 & 3.20 & 3.12  & 3.13
\\
\hline
\textcolor{black}{Proposed} filtering-out & \textbf{2.60} &	3.5 &	\textbf{3.03}	& 3.09	& \textbf{2.74}	& 3.25
\\
\hline

\end{tabular}
\end{table}

\paragraph{Computational time analysis}

It is reflected in Table \ref{tab:freq3} that, for lens blur (2.6s), S \& P (3.03s), and OE (2.74s), the filtering-out approach takes minimum computational time per image. Gaussian filters replace each pixel in an image with a weighted average of its neighbours based on a Gaussian distribution, whereas median filters replace each pixel in an image with the median value of its neighbours. To obtain the median value, neighbouring pixels must be sorted, and the Gaussian weighted average is required for Gaussian filters, which are computationally expensive, especially in datasets such as CURE-OR containing larger-sized images. In contrast, we are filtering-out poor-quality images in which the SSIM is below an optimum threshold value (0.9). In the above-mentioned impacts, many images \textcolor{black}{(79.2\% of historical dataset images, see Table \ref{tab:freq10})} are discarded based on the optimal threshold value of the SSIM. In other impacts such as CT, our proposed approach takes the second-lowest time. The lens dirty, and UE takes more computational time because the SSIM value is 0.1. \textcolor{black}{Since the filtering-out poor-quality images start from SSIM = 0.1 (see Step 8 of Algorithm 1), it appears that no images were filtered out (i.e., all historical images were kept). Consequentially, no images were filtered out during testing (i.e.,  while calculating recognition accuracy) which took more computational time as compared to the other impacts where a lot of images were discarded. \textcolor{black} {Table \ref{tab:freq10} } reflects the percentage of historical images left after filtering-out images based on the optimum threshold value of SSIM.}  

\subsection{Experimental results for CURE-TSR dataset}
This section provides the results for \textcolor{black}{optimum} threshold selection for the CURE-TSR dataset. An analysis of the performance of recognition accuracy and computational time is also discussed in this section.

\subsubsection{Estimating the \textcolor{black}{optimum} threshold for filtering-out poor-quality images for CURE-TSR dataset}
\begin{figure}
\includegraphics[width=3.5 in]{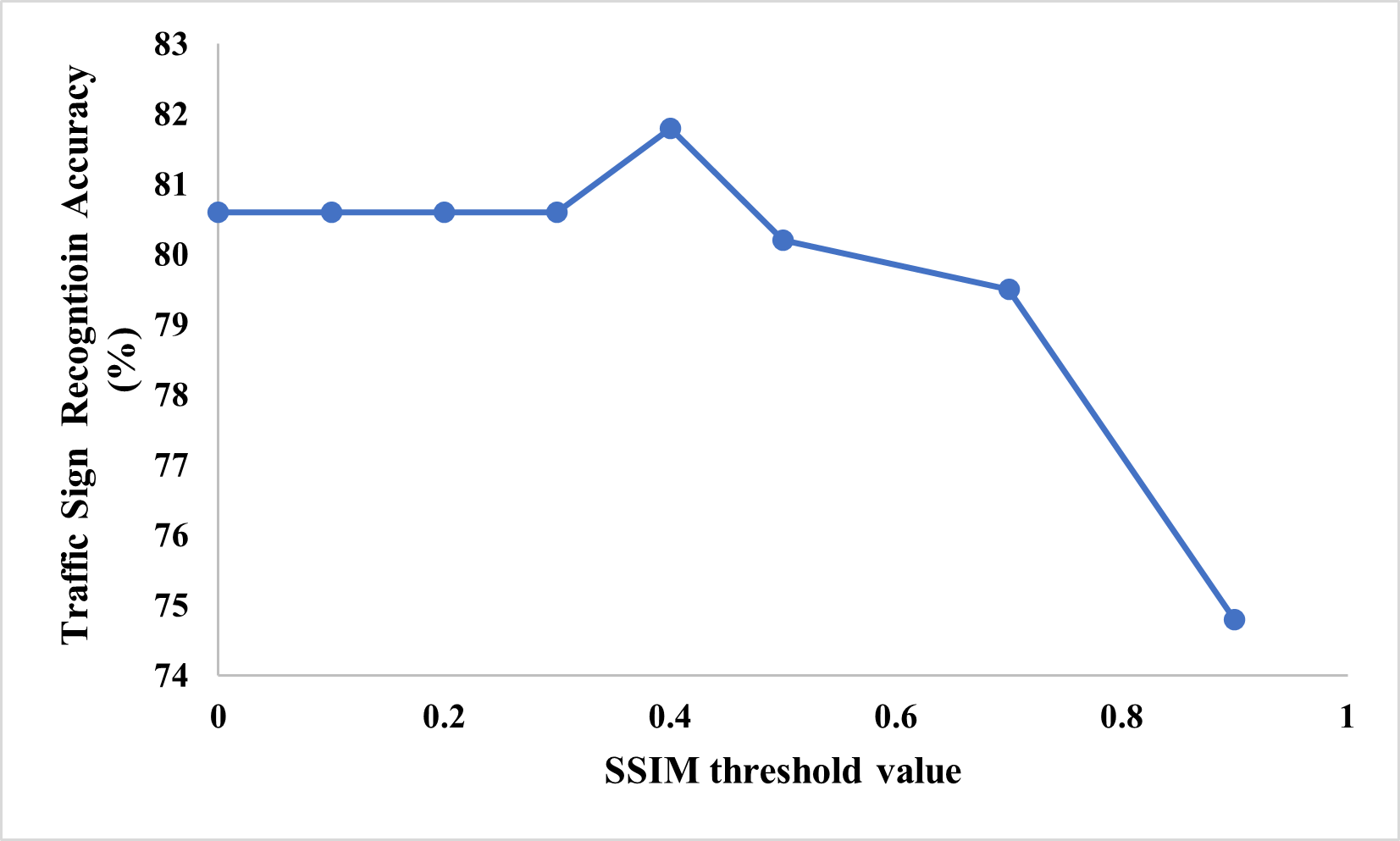}
\caption{SSIM threshold value vs object recognition accuracy for SH validation dataset of CURE-TSR.}\label{Shadow1}
\end{figure}

%\textcolor{black}{please change the description of the whole section}
Table \ref{tab:freq11} shows various impacts and their chosen SSIM threshold values for the CURE-TSR dataset. As shown in Figure \ref{Shadow1}, for SH, the maximum accuracy (81.8\%) is seen at SSIM value = 0.4.The fraction of remaining images is 20\% (refer to Table \ref{tab:freq11}).
For lens blur, the maximum accuracy (76.7\%) is at SSIM value = 0.5 \textcolor{black}{and the remaining fraction of images after filtering out poor-quality images is 20\%.  } 
%\textcolor{black}{be careful! The figure does not show maximum accuracy at SSIM=0.5}. 
Similar patterns are observed in lens dirty and it has maximum recognition accuracy (80.7\%) at SSIM value = 0.6 and \textcolor{black}{it has 80\% of remaining images after filtering-out poor-quality images}. For rain, maximum accuracy (78.7\%) is achieved at SSIM value = 0.1. 
%(refer to Figure \ref{Rain1}). 
Similarly, the maximum recognition accuracy (81.4\%) for snow is seen at SSIM value = 0.8, and the fraction of remaining images is 80\%. 
%as mentioned in Figures \ref{SH1} and \ref{Snow1}. 
Based on these selected threshold values, we analyse the performance of our CNN model which is covered in the following section.

\begin{table}[hbt!]
\caption{Various impacts in CURE-TSR and their optimum image quality (SSIM) threshold values. Note, that OSTV represents the optimum SSIM threshold value and RI is the remaining images.}
\label{tab:freq11}
\centering
\begin{tabular}{|c|c|c|c|c|c|}
\hline
{\textbf{Impact}}  & {\textbf{OSTV}} & {\textbf{Accuracy (\%)}} & {\textbf{RI (\%)}}\\
\hline
Lens blur & 0.5 & 76.7  &  20  \\
\hline
Lens dirty & 0.6 & 80.7  & 80\\
\hline
Rain & 0.1 & 78.7 & 100 \\
\hline
SH & 0.4 & 81.8 & 20  \\
\hline
Snow & 0.8 & 81.4 & 80\\
\hline
DK & 0.05 & 79.8 & 80 \\
\hline
\end{tabular}
\end{table}

\subsubsection{Performance analysis}
The following is an analysis of the performance of the CURE-TSR dataset in terms of recognition accuracy and computation time.
\paragraph{Recognition accuracy analysis}
The recognition accuracy of CURE-TSR, along with other approaches is revealed in Table \ref{tab:freq1}. Our proposed approach has maximum traffic sign recognition accuracy for impacts such as lens blur (91.3\%), lens dirty (94.3\%), snow (95.1\%), rain (92.6\%), and SH (95.3\%) in comparison with other approaches. 

\begin{table}[hbt!]
\caption{Traffic sign recognition \textcolor{black}{(TSR)} accuracy (\%) of CNN in different environmental and camera impacts with various denoising approaches for CURE-TSR. }
\label{tab:freq mean}
\centering
\begin{tabular}{|p{1.4cm}|p{0.5cm}|p{0.5cm}|p{0.5cm}|p{0.5cm}|p{0.5cm}|p{0.5cm}|p{0.5cm}|}
%\begin{tabular}{|c|c|c|c|c|c|c|c|c|c|c|c|c|c|}
\hline
& \multicolumn{7}{|c|}{\textbf{\textcolor{black}{TSR} accuracy for each impact (\%)}}   \\
 \hline
{\diagbox[width=\dimexpr \textwidth/11+1\tabcolsep\relax, height=0.8cm]{\textbf{Approach}}{\textbf{Impact}}}  & {\textbf{Blur} }& {\textbf{Dirty}} & { \textbf{Rain}} & {\textbf{Snow}} & {\textbf{SH}} & {\textbf{DK} } & {\textbf{Mean}} \\
\hline
\hline
Without Denoising  & 86.9 & 94.2 & \textbf{92.6} & 0.7  & 95 & \textbf{95.6} &  77.5	\\
\hline
Embedded median & 71.8 & 87.6 & 84.4 & 35.6 & 90.7 & 89.2 & 76.5\\
\hline
Embedded Gaussian & 71.3 & 88.1 & 86.3 & 1.9 & 91.1 & 89.2 & 71.3 \\
\hline
Traditional median & 73.3 & 85.3 & 79.5 & 35.6 & 87.8 & 84.8  & 74.3 \\
\hline
Traditional Gaussian & 70.3 & 80.4 & 74.9 & 1.9 & 83.3 & 1.9 & 52.1 \\
\hline
\textcolor{black}{Proposed} filtering-out & \textbf{91.3} & \textbf{94.3} & \textbf{92.6} & \textbf{95.1} &  \textbf{95.3} & 94.5 & \textbf{93.8} \\
\hline
\end{tabular}
\end{table}

In DK, our proposed approach has the second highest recognition accuracy, whereas the approach without denoising achieves the maximum recognition accuracy (i.e., 95.6\%).  Therefore, it is concluded that none of the discussed techniques works for denoising of DK. There is a need for image enhancement algorithms that can help to improve the quality and fidelity of dark regions in images, enhancing their overall visual impact and interpretability. \textcolor{black}{Besides this, our filtering-out approach produces the highest recognition accuracy for other impacts. For rain, along with the filtering approach, without filtering also achieves the highest accuracy (i.e., 92.6\%). Overall, the average accuracy of our proposed approach is 93.8\%, which is the highest when compared with other state-of-the-art approaches. The average improvement in accuracy achieved for almost all impacts vindicates the effectiveness of our filtering out approach.} 

\begin{table}[hbt!]
\caption{Computational time (\textcolor{black}{CPU time in} seconds) per image for traffic sign recognition in different impacts with various denoising approaches in CURE-TSR.}
\label{tab:freq2}
\centering
\begin{tabular}{|p{1.5cm}|p{0.5cm}|p{0.5cm}|p{0.5cm}|p{0.5cm}|p{0.5cm}|p{0.5cm}|p{0.5cm}|}
%\begin{tabular}{|c|c|c|c|c|c|c|c|c|c|c|c|c|c|}
\hline
& \multicolumn{6}{|c|}{\textbf{Computational time (seconds) for each impact}}  \\
 \hline
\diagbox[width=\dimexpr \textwidth/12+2\tabcolsep\relax, height=0.8cm]{\textbf{Approach}}{\textbf{Impact}} & {\textbf{Blur} }& {\textbf{Dirty}} & { \textbf{Rain}} & {\textbf{Snow}} & {\textbf{SH}} & {\textbf{DK}} \\
\hline
\hline
  Without Denoising  & 0.395 & 0.381 & 0.392 & 0.416 & 0.398 & 0.406\\
\hline
Embedded median & 0.363	& 0.378 &	0.391 &	0.374 &	0.375 & 0.398 
\\
\hline
Embedded Gaussian & \textbf{0.256} &	\textbf{0.265} & \textbf{0.262} &	\textbf{0.257} &	\textbf{0.277} &	\textbf{0.262} 
\\
\hline
 Traditional median & 0.332   & 0.369 &	0.391 &	0.311 &	0.321 & 0.310 
\\
\hline
Traditional Gaussian & 0.299	& 0.367 & 0.388 &	0.312 &	0.310 &	0.321 
\\
\hline
\textcolor{black}{Proposed} filtering-out & 0.362 &	0.328 &	0.329 & 0.311 &	0.310 &	0.278   \\
\hline
\end{tabular}
\end{table}

\paragraph{Computational time analysis}
Table \ref{tab:freq3} shows the computational time for all approaches including our proposed approach. Our proposed approach takes the \textcolor{black}{second lowest} computational time per image for impacts such as lens dirty (0.32s), rain (0.32s), snow (0.31s), SH (0.31s), DK (0.27s) in comparison with other approaches. \textcolor{black}{For lens blur, it takes 0.36s, which is third lowest computational time.}

The embedded Gaussian approach takes the least computational time for all impacts. A Gaussian filter is a linear operation that can be efficiently implemented using convolution operations, which are computationally fast, especially when the images are smaller in size and when optimized using libraries such as TensorFlow or PyTorch. Gaussian filtering is relatively simpler and more efficient than some non-linear denoising methods that involve complex operations such as thresholding or iterative optimization.
However, even though the embedded Gaussian approach may provide computational advantages, it has very less recognition accuracy for all impacts in comparison with our proposed filtering-out approach. For instance, the recognition accuracy of lens blur from embedded Gaussian filtering  is 71.3\%. In contrast, the recognition accuracy of lens blur in our proposed filtering out approach is 91.3\%. Therefore, slightly longer computation time is regarded as acceptable when the merits of improved accuracy for almost all impacts obtained using the filtering out approach are perceived. The embedded Gaussian approach does not always provide the best denoising performance, especially in scenarios with complex noise patterns or high noise levels. Because of this, it is important to consider both computational efficiency and denoising performance when choosing a denoising approach.

\section{Conclusions} \label{conclusions}

In this paper, we proposed an approach for filtering out poor-quality images by exploiting the assessed image quality. For evaluation, the SSIM was used to calculate the image quality because the SSIM is one of the most reliable image quality assessment metrics. 
We analyzed the performance without applying any denoising techniques, denoising with standard filters and CNN-based denoising techniques to prove the efficacy of our proposed approach. The performance of CNN was assessed in terms of recognition accuracy and computational time. 
Extensive experiments were conducted on two public datasets (CURE-TSR and CURE-OR). 
The experimental results demonstrate that our proposed filtering-out approach performs best in terms of recognition accuracy for most of the environmental and camera impacts in both CURE-TSR and CURE-OR datasets. In terms of computational time, our approach is comparable with other methods and it is lowest for lens blur, S \& P, and OE. \textcolor{black}{Even though it was difficult to get superior results for all environmental and camera impacts, the results show that our method has improved recognition accuracy for all impact types, except DK for the CURE-TSR dataset.} 
For DK, the without denoising approach achievesd 95.6\% recognition accuracy compared with 94.5\% accuracy obtained using our approach (see Table IV).
Conversely, we only worked with two datasets and two applications in this work. The use of other image-recognition benchmark datasets for different applications could be the future direction that makes filtering-out approach applicable to other real-world applications involving different types of image analysis, including image segmentation and recognition such as surveillance and security, forensic analysis and remote sensing.

\ifCLASSOPTIONcaptionsoff
  \newpage
\fi

\addcontentsline{toc}{section}{References}
\bibliographystyle{IEEEtran}
\bibliography{thebibliography}

\vspace{-2cm}  
\begin{IEEEbiography}
[{\includegraphics[width=1in,height=1.2in]{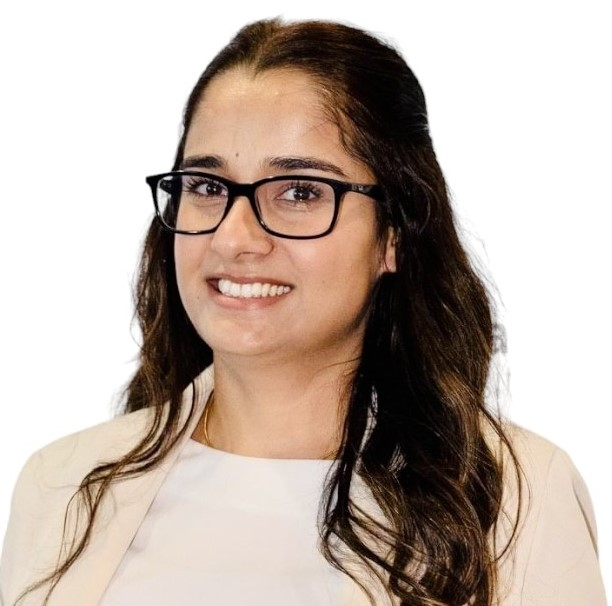}}]{Roopdeep Kaur} received a B.Tech in Electronics and Communication from Punjab Technical University in 2016, an M.E. from Thapar University in 2018, and a Ph.D. from Federation University Australia in 2024. Her research focuses on the Internet of Things, image processing, and data analytics.
\end{IEEEbiography}

\vspace{-1cm}
\begin{IEEEbiography}[{\includegraphics[width=1in,height=1.2in]{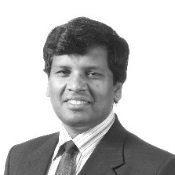}}]{Gour Karmakar}
(M’01) received his Ph.D. in IT from Monash University in 2003 and is currently a Professor at Federation University Australia. With over 193 peer-reviewed publications, including 52 in reputed journals, his research is supported by industry and competitive grants, including an ARC Linkage Grant in 2011. His interests include multimedia signal processing, AI, IoT, and cybersecurity.
\end{IEEEbiography}

\vspace{-16.5cm}
\begin{IEEEbiography}
[{\includegraphics[width=1in,height=1.2in]{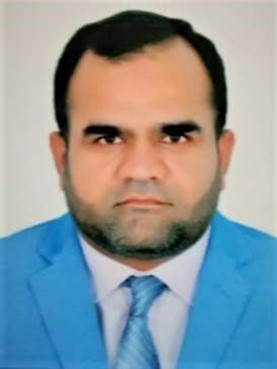}}]
{Muhammad Imran} is a Senior Lecturer at the Institute of Innovation, Science, and Sustainability, Federation University Australia. His research focuses on mobile and wireless networks, IoT, big data analytics, cloud/edge computing, and information security. He leads the Wireless Networks and Security (WINS) research group and has collaborated on numerous international research projects. With over 300 publications in leading conferences and journals, he was recognized as a 2022 Clarivate Highly Cited Researcher in Computer Science. He has held editorial roles in prestigious journals, including IEEE Network, FGCS, and IEEE Communications Magazine, and has been named an Outstanding Associate Editor multiple times.

\end{IEEEbiography}
\end{document}